\documentclass[10pt,twocolumn,letterpaper]{article}
\usepackage[accsupp]{axessibility}  

\usepackage{wacv}
\usepackage{times}
\usepackage{epsfig}
\usepackage{graphicx}
\usepackage{amsmath}
\usepackage{amssymb}
\usepackage{booktabs}
\usepackage{multicol}
\usepackage{subfigure}
\usepackage{multirow}
\usepackage{color, colortbl}
\definecolor{Gray}{gray}{0.9}

\def\wacvPaperID{2069} 

\wacvalgorithmstrack   

\wacvfinalcopy 

\ifwacvfinal
\usepackage[breaklinks=true,bookmarks=false]{hyperref}
\else
\usepackage[pagebackref=true,breaklinks=true,colorlinks,bookmarks=false]{hyperref}
\fi

\begin{document}

\title{Elimination of Non-Novel Segments at Multi-Scale for Few-Shot Segmentation}

\author{Alper Kayabaşı$^{1, 2}$,\; Gülin Tüfekci$^{1, 2}$,\; İlkay Ulusoy$^{2}$ \\
$^{1}$Research Center, Aselsan Inc \\
$^{2}$Middle East Technical University, Ankara, Turkey \\
{\tt\small \{alper.kayabasi,gulin.tufekci,ilkay\}@metu.edu.tr}
}


\maketitle
\thispagestyle{empty}

\begin{abstract}
   Few-shot segmentation aims to devise a generalizing model that segments query images from unseen classes during training with the guidance of a few support images whose class tally with the class of the query. There exist two domain-specific problems mentioned in the previous works, namely spatial inconsistency and bias towards seen classes. Taking the former problem into account, our method compares the support feature map with the query feature map at multi scales to become scale-agnostic. As a solution to the latter problem, a supervised model, called as base learner, is trained on available classes to accurately identify pixels belonging to seen classes. Hence, subsequent meta learner has a chance to discard areas belonging to seen classes with the help of an ensemble learning model that coordinates meta learner with the base learner. We simultaneously address these two vital problems for the first time and achieve state-of-the-art performances on both PASCAL-5\textsuperscript{i} and COCO-20\textsuperscript{i} datasets.
\end{abstract}

\section{Introduction} 

Semantic segmentation is a crucial task that classifies each pixel of an image to make sense of the scene with application areas such as autonomous driving \cite{auto_app} and medical imaging \cite{unet}. Deep learning pervades semantic segmentation like other tasks of computer vision \cite{deeplab1, fcn}. Supervised segmentation models are required to employ abundant annotated data belonging to each class in the training set since the generalization capacity of supervised models decreases with scarce labeled data. Therefore, adapting the model to work on unseen classes requires dense annotation of myriad data from novel classes. Shaban \etal \cite{shaban} proposed few-shot segmentation to remove the labeling effort and increase the generalization capacity of a model given few data for the first time. 

\begin{figure}[t]
\centering
\begin{tabular}{c}
    \includegraphics[width=0.9\linewidth]{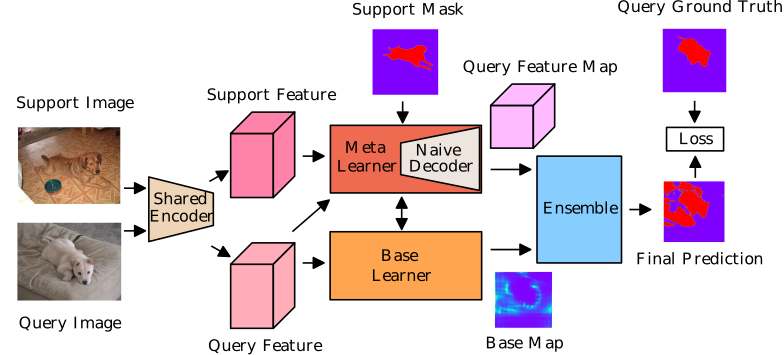} \\
    (a) \\
    \includegraphics[width=0.9\linewidth]{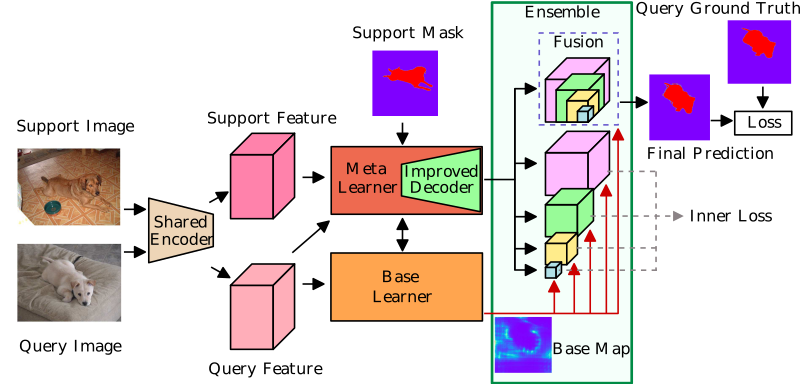} \\
    (b)
\end{tabular}
\caption{(a) Overview of BAM \cite{bam}. Support and query features are used by meta learner to extract support feature map while base learner provides guidance for the base classes and leads the meta learner to focus on novel regions via ensembling. (b) Our proposed method. The decoder for meta learner is improved such that query feature map is obtained at multi-scale. Support feature map is compared with query feature maps at multi-scale to obtain enriched query features. Query predictions obtained from enriched query features at each scale are ensembled with the base map as well as the prediction obtained from the fusion of them. Inner losses are computed at different scale levels and the final prediction is obtained from the ensemble of the base map with the predictions from the fused query feature maps. (Best viewed in zoom)}
\label{fig:fig1}
\end{figure}

Few-shot segmentation addresses the problem of making pixel-wise predictions for a target image, called a query, from an unseen class with the guidance of a support image from the same category. Inspired by the few-shot classification task \cite{rise_of_episode}, most methods utilized the episodic training strategy in which the gradients are averaged over tasks named as an episode. Each episode is sampled from a dataset whose classes are disjoint from a test case where only a few data are available. These episodes are used to imitate the test case during training to prevent overfitting. Despite this intention, a model trained with this strategy tends to mistake segments from seen training classes, referred as \textit{base classes}, as \textit{novel classes} because of constantly experiencing the same set of classes during training. Hence, the co-occurrence of novel and base classes in the same scene causes entanglement between features of pixels that are part of the novel and base categories.

To prevent this entanglement, prediction of the supervised model which is trained on base classes guides the meta learner, which is responsible for detecting novel areas. Meta learner is directed to areas not occupied by the base classes so that contradiction between the base learner and the meta learner is avoided via the ensemble model entailing both base and meta predictions \cite{bam}. On the side of meta learner, candidate objects in query image might not cover as same area as those in support images; so, the model should compare support feature map with query feature map at different resolutions to disentangle adjacent regions around novel segments \cite{pfenet}. As shown in Fig. \ref{fig:fig1}, the ensemble of base and meta learner without improved decoder fails to distinguish background from foreground since naive decoder, which is designed for the supervised scheme, lacks to combine features at different resolutions in favor of complete query prediction. Hence, we transform the naive decoder into an improved decoder such that not only does it correlate the support image with the query image at multi-resolution but also it benefits from merits of base learner at multi-resolution. In this regard, we hypothesize that there are cases where it is not enough that base learner discourages meta learner from base regions at single-scale. Our experiments verify that the improved decoder and ensembling the predictions at multi-scale outperform the decoder equipped with ensembling the prediction at single-scale. Our contributions in this paper are two-fold:

\begin{itemize}
    \item We alleviate the spatial inconsistency and the bias problems together with the assistance of our proposed decoder that seeks to remove bias at multi-resolution.
    \item Our proposed method achieves new state-of-the art performance on both PASCAL-5$^i$ (mIoU @ 1-shot: 68.59\%, mIoU @ 5-shot: 72.05\%) and COCO-20$^i$ (mIoU @ 1-shot: 47.16\%, mIoU @ 5-shot: 52.50\%) datasets for few-shot segmentation task.
\end{itemize}
  
\section{Related work} 

\subsection{Semantic segmentation} 

Fully convolutional neural network (FCN), which is the pioneering work in semantic segmentation field, formulates semantic segmentation as a pixel-wise classification task \cite{fcn}. In FCN, all fully connected layers at the end of a model are transformed into convolution layers so that the network accepts arbitrary input sizes. Success of FCN accelerates the field and results in outstanding architectures such as UNet \cite{unet}, PSPNet \cite{pspnet}, and Deeplab \cite{deeplab1, deeplab2}. PSPNet combines average pooled feature maps at different scales to contain not only global but also local context \cite{pspnet}. Deeplab introduces ASPP module \cite{deeplab1} equipped with dilated convolution that increases the receptive field of the network without a decrease in resolution by inserting holes between filter weights. 

\subsection{Few-shot segmentation} 

Few-shot segmentation studies can be categorized into four groups according to their objectives: imbalance in details, inter-class gap, spatial inconsistency, and correlation reliability. 

The imbalance in details problem emphasizes that there might be details that do not co-exist in both query and corresponding support image. Hence, inconsistent regions between support and query should be detected and eliminated on the support side to prevent redundant details or noise in an adaptive manner. PGNet \cite{pgnet} proposes a network that associates each query pixel to the relevant parts of the support image to remove noise, where relevancy is generally quantified by a similarity metric such as cosine similarity. PANet \cite{panet} adds regularization loss to ensure that the network becomes successful if the roles of support and query are swapped. ASGNet \cite{asgnet} aims to find an adaptive number of prototypes and their spatial extents based on image content with a boundary-aware superpixel algorithm so that prototypes represent parts of an object with similar characteristics. Each query pixel utilizes support prototype giving maximum cosine similarity with itself as reference.  

Most approaches assume that transferable knowledge constantly exists in base set and tries to segment image from unseen classes during training. This strong assumption loses its validity in proportion to the discrepancy between base and novel dataset, and this problem is referred as an inter-class gap. RePRI \cite{repri} shows that the severity of overfitting is exaggerated in few-shot segmentation and adapting novel classes by fine-tuning over support images improves segmentation performance. CWT \cite{cwt} episodically trains self-attention block that adapts classifier weights of network updated over support image during both training and test. BriNet \cite{brinet} regards prediction for query as pseudo-mask and switches the roles between query and support to update model with ground truth support mask during test stage until determined mIoU threshold is exceeded for support mask and its prediction. 

Architectures designed for supervised cases fail to provide scale-invariance in few-shot scenarios since contextual relationships are not figured out by a handful of data. Methods design control mechanism that provides favorable information exchange between different resolutions \cite{sagnn, cyclic_mem, pfenet, pgnet}.

Correlation map determines pixel-wise similarity between support and query images. Problems such as background clutter and occlusion lead to noise in the correlation map; hence, it results in erroneous comparisons and training processes based on misinterpreted correspondences, which is called as a hyper correlation reliability problem. Methods are proposed to check the validity of correspondences based on learnable or engineered criteria so that filtered correlation maps become interpretable. After the elimination of deceptive correspondences, all similarities corresponding to each query pixel from the support image are summed to obtain an activation score that determines the level of association of that query pixel with the foreground of support \cite{hsnet, cat}.

\section{Preliminaries} 

In order to relate the specified challenges in few-shot segmentation with our approach and guide the reader through steps, the task is formally defined in Section \ref{task_description}, Feature Enrichment Module (FEM) \cite{pfenet} and Base and Meta Learner (BAM) \cite{bam} are introduced in Sections \ref{revisit_fem} and \ref{revisit_bam}, respectively.

\subsection{Task description} \label{task_description}

Few-shot segmentation task utilizes base dataset containing adequate images with their annotations whose classes, $\mathcal{C}_{base}$, are disjoint from novel classes, $\mathcal{C}_{novel}$, in which dense predictions are fulfilled with few data and their annotations. K number of available data and their annotations belonging to the novel classes constitute a support set $\mathcal{S}$ for testing which are expected to guide a model ${M}$ to make predictions for query image $\mathcal{I}_q$, which is dubbed as K-shot segmentation. The support set is formally represented as $\mathcal{S}=\{\mathcal{I}_{s_i}, \mathcal{M}_{s_i}\}_{i=1}^{K}$, where $\mathcal{I}_{s_i}$, and $\mathcal{M}_{s_i}$ correspond to i$^\text{th}$ support image and its dense ground truth mask. On the side of training, support set for training is sampled from base dataset along with query set which consists of the query image and its ground truth, sharing its class with the chosen support set. The aforementioned classes are treated as novel class during training in order to perform episodic training, where pixels belonging to chosen class are assigned as foreground while pixels from all other classes are considered as background. Query set is formally represented as $\mathcal{Q}=\{\mathcal{I}_{q}, \mathcal{M}_{q}\}$, where $\mathcal{I}_{q}$ and $\mathcal{M}_{q}$ correspond to the query image and its dense ground truth mask. The model, ${M}$, is trained by backpropagating binary cross entropy loss between $\mathcal{I}_{q}$ and $\mathcal{M}_{q}$ over tasks, named as episodes involving the selected support set from base dataset with the accompanying query set.

\subsection{Revisit feature enrichment module} \label{revisit_fem} 

Multi-scale modules in supervised semantic segmentation generally do not provide mechanism to form independent interaction between masked global average pooled support feature map, called as support prototype, and average pooled query feature map at different scales. For example, conventional multi-scale architectures apply single filtering to combination of query feature map, support prototype, and prior mask that describes likelihood of query pixel being related with at least one pixel in foreground of support \cite{pfenet}. Different from these approaches, inter-source enrichment module of FEM separately applies the filtering to the query feature map at each different scale, which is combined with support prototype and prior mask. Furthermore, inter-scale interaction module of FEM fulfills the information transfer between two consecutive resolutions in top-down path, where top-down path consists of outputs of inter-source enrichment module ordered from high resolution to low resolution. During information transfer, preservation of hierarchical structure allows gradual accumulation of information from higher resolution to lower resolution. In this module, each resolution has direct connection only to its neighbour in the top-down direction. Therefore, there is no connection between any resolution pairs other than the consecutive ones. Hence, the module has a chance to decide on the scale at which the obtained information is sufficient to make a prediction and the following scales would bring redundancy. Following this reasoning, feature maps at different resolutions are fused via information concentration module in FEM. 

\subsection{Revisit base and meta learner} \label{revisit_bam} 

Typical few-shot segmentation approaches use meta-learning approach such that the knowledge gained from training the model on the base classes is utilized to predict the mask of the query image belonging to a novel class given a support image belonging to the same novel class. This process is called as meta-learning since learning tasks are sampled from the base classes during training in order to simulate the few-shot settings in testing so that the training and testing conditions are matched. However, as \cite{bam} states, training on base classes introduce a bias towards them during testing, which prevents the model to work on the novel classes properly. To tackle this bias BAM is introduced, where a base learner, apart from the meta learner, explicitly works on the known classes. When the information related to known classes is used during testing, the recognition of novel classes would be enhanced.

Training BAM consists of two stages, namely base-training and meta-training. Both learners share the same backbone as feature encoder. To leverage the representations at different levels of abstraction, features are obtained from different layers of the encoder. Base learner is trained in a supervised manner so that the ability to make confident predictions regarding base classes is gained. In meta-training stage, the parameters of the base learner are fixed. The features of support and query images are extracted by the shared encoder and the features obtained after block-2 and block-3 of ResNet-50 \cite{resnet} are concatenated and transformed with 1$\times$1 convolution layer, which are denoted by $\textbf{f}_m^s$ and $\textbf{f}_m^q$ respectively. Query features after block-4 of ResNet-50, $\textbf{f}_b^q$, are processed by base learner and decoded by Pyramid Scene Parsing Network (PSPNet) \cite{pspnet}, which is composed of Pyramid Pooling Module (PPM) and classifier so that the probability map of base classes, $\textbf{p}_b^f$, is obtained. This step is crucial since the base classes are the background classes for the query image while the novel class is the foreground, which is to be predicted by the meta learner. The support mask, $\textbf{m}^s$, is used together with $\textbf{f}_m^s$ in order to obtain the support prototype, $\textbf{v}_s$. The query features, support prototype and the prior map are concatenated, which is inputted to the meta decoder. At the end of the meta decoder, output background and foreground probability maps, $\textbf{p}_m^0$ and $\textbf{p}_m^1$, are obtained.

Low level features for support and query images are obtained from the intermediate levels of the encoder, denoted by $\textbf{f}_{low}^s$ and $\textbf{f}_{low}^q$, and the Frobenius norm between their Gram matrices is computed as adjustment factor, $\psi$. The adjustment factor is leveraged such that the smaller $\psi$ is, the closer the representations of support and query images become. In other words, as $\psi$ gets smaller, the reliability of the prediction of the meta learner increases such that the query features become representative of the support features. Moreover, $\textbf{p}_m^0$ is ensembled with $\textbf{p}_b^f$ in order to force the pixels belonging to non-novel regions for the query image to be closer to the base classes. This enhanced information is used such that the corresponding pixels are less likely to be predicted as novel. Resultant ensembled information is concatenated with $\textbf{p}_m^1$ in order to produce final prediction.

\section{Method}

As CANet implies \cite{canet}, we use the middle level features by applying 1x1 convolution to concatenation of feature maps obtained from block-2 and block-3. We represent middle level features belonging to support and querys image respectively as in Eq. \ref{eq:supp_feat} and Eq. \ref{eq:query_feat}, where $Enc$ symbolizes the middle level feature extractor.

\begin{equation}
\label{eq:supp_feat}
    \textbf{f}_m^s = Enc(\mathcal{I}_s)\; \in R^{H \times W \times C}
\end{equation}

\begin{equation}
\label{eq:query_feat}
    \textbf{f}_m^q = Enc(\mathcal{I}_q)\; \in R^{H \times W \times C}
\end{equation}

To obtain the prior map in a similar manner to PFENet \cite{pfenet}, high level query and support features are reshaped from R$^{H \times W \times  C}$ to R$^{HW \times C}$ at first. After that, row wise norms for high level query and support pixel features are computed respectively as in Eq. \ref{eq:norm_fq} and Eq. \ref{eq:norm_fs}, where $^\circ$ corresponds to Hadamard root while $diag$ outputs diagonal elements of a matrix as a column vector. 

\begin{equation}
\label{eq:norm_fq}
    \lVert \tilde{\textbf{f}_b^q} \rVert =  (diag(\tilde{\textbf{f}_b^q} \times \tilde{\textbf{f}_b^q}^\intercal))^{\circ{1/2}} \in R^{HW \times 1}
\end{equation}
\begin{equation}
\label{eq:norm_fs}
        \lVert \tilde{\textbf{f}_b^s} \rVert =  (diag(\tilde{\textbf{f}_b^s} \times \tilde{\textbf{f}_b^s}^\intercal))^{\circ{1/2}} \in R^{HW \times 1}
\end{equation}

Prior map is calculated by max pooling the cosine similarity matrix between the high level query and support pixels along row wise direction as shown in Eq. \ref{eq:prior_map}, where $\oslash$ is Hadamard divison.

\begin{equation}
\label{eq:prior_map}
    \textbf{C}_q = \text{pool}((\tilde{\textbf{f}_b^q}  \times \tilde{\textbf{f}_b^s}^\intercal) \oslash (\lVert \tilde{\textbf{f}_b^q} \rVert \times \lVert \tilde{\textbf{f}_b^s} \rVert^\intercal))\; \in R^{H \times W \times 1}
\end{equation}

Masked global average pooling is applied to $\textbf{f}_m^s$ to extract support prototype, $\textbf{v}_s$, in Eq. \ref{eq:prototype}, where $\mathcal{R}$ downsamples $\mathcal{M}_s$ to the size of $\textbf{f}_m^s$.

\begin{equation}
\label{eq:prototype}
    \textbf{v}_s = \text{masked\_avg\_pool}(\textbf{f}_m^s,\mathcal{R}\left(\mathcal{M}_s\right))\; \in R^{1 \times 1 \times C}
\end{equation}

$FEM$ takes $\textbf{v}_s, \textbf{C}_q$, and $\textbf{f}_m^{q}$ as input and outputs N+1 enriched query feature maps where N of them correspond to enriched auxiliary feature maps at N different scales and the last one is the fusion of them as shown in Eq. \ref{eq:fem}.

\begin{equation}
\label{eq:fem}
    \textbf{X}_q^{s_1}, \textbf{X}_q^{s_2}, ... , \textbf{X}_q^{s_N}, \textbf{X}_q^{fused} = FEM(\textbf{C}_q, \textbf{f}_m^q, \textbf{v}_s)
\end{equation}

\begin{equation}
\label{eq:aux_classifier}
    \textbf{C}^{AUX} = \{\textbf{C}^{aux,1}, \textbf{C}^{aux,2}, ..., \textbf{C}^{aux,N}, \textbf{C}^{aux,fused}\}
\end{equation}

$\textbf{C}^{AUX}$ in Eq. \ref{eq:aux_classifier} represents set of classifiers, where first N classifiers correspond to auxiliary classifiers, which make predictions for the multi-scale features, while the last classifier is responsible for the prediction deduced from the fused feature. By using these classifiers, we obtain background and foreground logit values for enriched query feature maps at each scale and the fused feature map respectively in Eq. \ref{eq:feature_to_prob_map} and Eq. \ref{eq:feature_to_prob_map_fused}, where $\oplus$ performs concatenation operation.

\begin{equation}
\label{eq:feature_to_prob_map}
    \textbf{p}_{m,s_i}^0, \textbf{p}_{m,s_i}^1 = \textbf{C}^{aux,i}(\textbf{X}_q^{s_i})
\end{equation}

\begin{equation}
\label{eq:p_msi}
    \textbf{p}_{m,s_i} = \textbf{p}_{m,s_i}^0 \oplus \textbf{p}_{m,s_i}^1
\end{equation}

\begin{equation}
\label{eq:feature_to_prob_map_fused}
    \textbf{p}_{m,fused}^0, \textbf{p}_{m,fused}^1 = \textbf{C}^{aux,fused}(\textbf{X}_q^{fused})
\end{equation}

\begin{equation}
\label{eq:p_mfused}
    \textbf{p}_{m,fused} = \textbf{p}_{m,fused}^0 \oplus \textbf{p}_{m,fused}^1
\end{equation}

\begin{figure*}[!h]
\centering
\resizebox{\linewidth}{!}{
\includegraphics[width=0.9\linewidth]{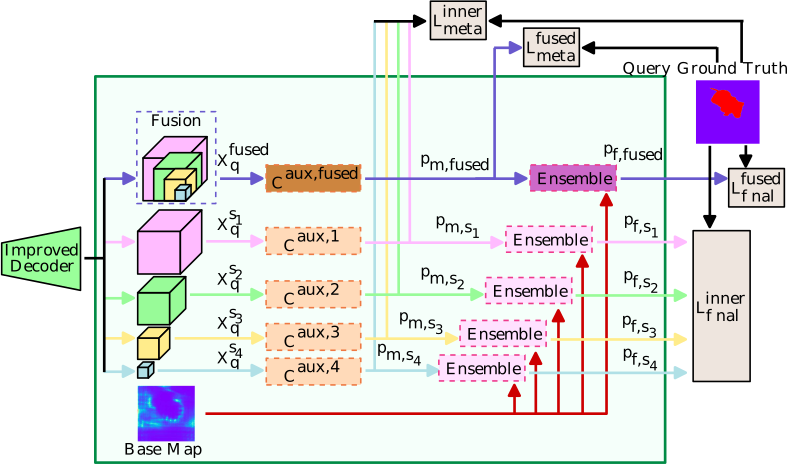}}
\caption{Detailed architecture of the multi-scale ensemble module. Features at multi-scale and the fusion of them are obtained at the end of the improved decoder as $\textbf{X}_q^{s_i}$ and $\textbf{X}_q^{fused}$ respectively, which are used by the corresponding auxiliary classifiers. The resultant enriched query feature maps are ensembled with the base map to obtain query predictions at multi-scale, which are denoted by $\textbf{p}_{f,s_i}$ and $\textbf{p}_{f,fused}$ respectively. Inner losses are computed from probability maps at intermediate scales $\left(\textbf{p}_{m,s_i}\right)$ and predictions at intermediate scales $\left(\textbf{p}_{f,s_i}\right)$ while fused losses are computed from fused probability maps $\left(\textbf{p}_{m,fused}\right)$ and fused predictions $\left(\textbf{p}_{f,fused}\right)$. (Best viewed in color)}
\label{fig:ensemble}
\end{figure*}

$BaseLearner$ in Eq. \ref{eq:prob_map_base} takes $\textbf{f}_q^m$ as input and outputs summation of predicted probabilities for all classes except background. 

\begin{equation}
\label{eq:prob_map_base}
    \textbf{p}_b^f = BaseLearner(\textbf{f}_m^q)
\end{equation}

While Eq. \ref{eq:ensemble_bg_fused}, Eq. \ref{eq:ensemble_fg_fused}, and Eq.  \ref{eq:p_ffused} are the same as in BAM \cite{bam}, we employ separate ensemble models for each auxiliary predictions to render meta-model aware of non-novel regions at each scale, as inspired by BAM, as shown in Eq. \ref{eq:ensemble_bg}, Eq. \ref{eq:ensemble_fg}, and Eq. \ref{eq:p_fsi} as well as in Fig. \ref{fig:ensemble} with pink rectangular boxes covered by dashed lines.

\begin{equation}
\label{eq:ensemble_bg_fused}
    \textbf{p}_{f,fused}^0 = Ens_\phi(\textbf{p}_b^f, Ens_\psi(\textbf{p}_{m,fused}^0, \psi))
\end{equation}

\begin{equation}
\label{eq:ensemble_fg_fused}
    \textbf{p}_{f,fused}^1 = Ens_\psi(\textbf{p}_{m,fused}^1, \psi)
\end{equation}

\begin{equation}
\label{eq:p_ffused}
    \textbf{p}_{f,fused} = \textbf{p}_{f,fused}^0 \oplus \textbf{p}_{f,fused}^1
\end{equation}

\begin{equation}
\label{eq:ensemble_bg}
    \textbf{p}_{f,s_i}^0 = Ens_{\phi,s_i}(\textbf{p}_b^f, Ens_\psi(\textbf{p}_{m,s_i}^0, \psi))
\end{equation}

\begin{equation}
\label{eq:ensemble_fg}
    \textbf{p}_{f,s_i}^1 = Ens_\psi(\textbf{p}_{m,s_i}^1, \psi)
\end{equation}

\begin{equation}
\label{eq:p_fsi}
    \textbf{p}_{f,s_i} = \textbf{p}_{f,s_i}^0 \oplus \textbf{p}_{f,s_i}^1
\end{equation}

\begin{equation}
\label{eq:loss_meta_inner}
    \mathcal{L}_{meta}^{inner} = \sum_{i=1}^{N}CE(\textbf{p}_{m,s_i}, \mathcal{M}_q)
\end{equation}

\begin{equation}
\label{eq:loss_meta_fused}
    \mathcal{L}_{meta}^{fused} = CE(\textbf{p}_{m,fused}, \mathcal{M}_q)
\end{equation}

\begin{equation}
\label{eq:loss_final_inner}
    \mathcal{L}_{final}^{inner} = \sum_{i=1}^{N}CE(\textbf{p}_{f,s_i}, \mathcal{M}_q)
\end{equation}

\begin{equation}
\label{eq:loss_final_fused}
    \mathcal{L}_{final}^{fused} = CE(\textbf{p}_{f,fused}, \mathcal{M}_q)
\end{equation}

Eq. \ref{eq:loss_meta_inner} and Eq. \ref{eq:loss_meta_fused} compute cross entropy losses for the auxiliary predictions and the fused prediction before ensemble respectively while Eq. \ref{eq:loss_final_inner} and Eq. \ref{eq:loss_final_fused} compute cross entropy losses for the auxiliary predictions and the fused prediction after ensemble accordingly.

\begin{equation}
\label{eq:loss}
    \mathcal{L}_{final}^{total} = \mathcal{L}_{meta}^{inner} + \mathcal{L}_{meta}^{fused} + \mathcal{L}_{final}^{inner} + \mathcal{L}_{final}^{fused}
\end{equation}

All individuals losses are accumulated to update the network eventually as in Eq. \ref{eq:loss}.

\begin{table*}[!h]
\centering
\resizebox{\linewidth}{!}{
\begin{tabular}{|c|c|c c c c c|c c c c c|}
\hline
\multirow{2}{*}{Backbone} & \multirow{2}{*}{Method} & \multicolumn{5}{|c|}{1-shot (\%)} & \multicolumn{5}{c|}{5-shot (\%)}\\
\cline{3-12}
 & & Fold-0 & Fold-1 & Fold-2 & Fold-3 & Average & Fold-0 & Fold-1 & Fold-2 & Fold-3 & Average\\
\hline
\multirow{5}{*}{VGG-16} & PFENet (TPAMI'20) \cite{pfenet} & 56.90 & 68.20 & 54.40 & 52.40 & 58.00 & 59.00 & 69.10 & 54.80 & 52.90 & 59.00\\
 & NTRENet (CVPR'22) \cite{ntrenet} & 57.70 & 67.60 & 57.10 & 53.70 & 59.00 & 60.30 & 68.00 & 55.20 & 57.10 & 60.20\\
 & DPCN (CVPR'22) \cite{dpcn} & 58.90 & 69.10 & 63.20 & 55.70 & 61.70 & 63.40 & 70.70 & 68.10 & 59.00 & 65.30\\
 & BAM (CVPR'22) \cite{bam} & \underline{63.18} & \underline{70.77} & \underline{66.14} & \underline{57.53} & \underline{64.41} & \underline{67.36} & \underline{73.05} & \underline{70.61} & \underline{64.00} & \underline{68.76}\\
 & \cellcolor{Gray}BAM++ (ours) & \cellcolor{Gray}\textbf{64.67} & \cellcolor{Gray}\textbf{72.11} & \cellcolor{Gray}\textbf{67.83} & \cellcolor{Gray}\textbf{59.47} & \cellcolor{Gray}\textbf{66.02} & \cellcolor{Gray}\textbf{69.40} & \cellcolor{Gray}\textbf{74.35} & \cellcolor{Gray}\textbf{72.77} & \cellcolor{Gray}\textbf{67.19} & \cellcolor{Gray}\textbf{70.93}\\
\hline
\multirow{7}{*}{ResNet-50} & PGNet (ICCV'19) \cite{pgnet} & 56.00 & 66.90 & 50.60 & 50.40 & 56.00 & 57.70 & 68.70 & 52.90 & 54.60 & 58.50\\
 & PFENet (TPAMI'20) \cite{pfenet} & 61.70 & 69.50 & 55.40 & 56.30 & 60.80 & 63.10 & 70.70 & 55.80 & 57.90 & 61.90\\
 & NTRENet (CVPR'22) \cite{ntrenet} & 65.40 & 72.30 & 59.40 & 59.80 & 64.20 & 66.20 & 72.80 & 61.70 & 62.20 & 65.70\\
 & ASNet (CVPR'22) \cite{asnet} & 68.90 & 71.70 & 61.10 & \underline{\textbf{62.70}} & 66.10 & \underline{\textbf{72.60}} & 74.30 & 65.30 & 67.10 & 70.80\\
 & DPCN (CVPR'22) \cite{dpcn} & 65.70 & 71.60 & \underline{69.10} & 60.60 & 66.70 & 70.00 & 73.20 & \underline{70.90} & 65.50 & 69.90\\
 & BAM (CVPR'22) \cite{bam} & \underline{68.97} & \underline{73.59} & 67.55 & 61.13 & \underline{67.81} & 70.59 & \underline{75.05} & 70.79 & \underline{67.20} & \underline{70.91}\\
 & \cellcolor{Gray}BAM++ (ours) & \cellcolor{Gray}\textbf{69.46} & \cellcolor{Gray}\textbf{74.16} & \cellcolor{Gray}\textbf{69.20} & \cellcolor{Gray}61.54 & \cellcolor{Gray}\textbf{68.59} & \cellcolor{Gray}70.81 & \cellcolor{Gray}\textbf{75.34} & \cellcolor{Gray}\textbf{73.04} & \cellcolor{Gray}\textbf{68.99} & \cellcolor{Gray}\textbf{72.05}\\
\hline
\end{tabular}}
\caption{1-shot and 5-shot class mIoU results on PASCAL-5$^i$ dataset for VGG-16 and ResNet-50 as backbone, provided for 4 folds and the average. The best results are given in \textbf{boldface}. The \underline{underlined} results show the best performance excluding our method.}
\label{table:results_pascal}
\end{table*}

\begin{table*}[!h]
\centering
\resizebox{\linewidth}{!}{
\begin{tabular}{|c|c|c c c c c|c c c c c|}
\hline
\multirow{2}{*}{Backbone} & \multirow{2}{*}{Method} & \multicolumn{5}{|c|}{1-shot (\%)} & \multicolumn{5}{c|}{5-shot (\%)}\\
\cline{3-12}
 & & Fold-0 & Fold-1 & Fold-2 & Fold-3 & Average & Fold-0 & Fold-1 & Fold-2 & Fold-3 & Average\\
\hline
\multirow{5}{*}{ResNet-50} & NTRENet (CVPR'22) \cite{ntrenet} & 36.80 & 42.60 & 39.90 & 37.90 & 39.30 & 38.20 & 44.10 & 40.40 & 38.40 & 40.30\\
 & ASNet (CVPR'22) \cite{asnet} & - & - & - & - & 42.20 & - & - & - & - & 68.80\\
 & DPCN (CVPR'22) \cite{dpcn} & 42.00 & 47.00 & 43.20 & 39.70 & 43.00 & 46.00 & \underline{54.90} & 50.80 & 47.40 & 49.80\\
 & BAM (CVPR'22) \cite{bam} & \underline{43.41} & \underline{50.59} & \underline{\textbf{47.49}} & \underline{43.42} & \underline{46.23} & \underline{49.26} & 54.20 & \underline{\textbf{51.63}} & \underline{\textbf{49.55}} & \underline{51.16}\\
 & \cellcolor{Gray}BAM++ (ours) & \cellcolor{Gray}\textbf{44.43} & \cellcolor{Gray}\textbf{51.98} & \cellcolor{Gray}47.01 & \cellcolor{Gray}\textbf{45.22} & \cellcolor{Gray}\textbf{47.16} & \cellcolor{Gray}\textbf{52.53} & \cellcolor{Gray}\textbf{57.02} & \cellcolor{Gray}50.97 & \cellcolor{Gray}49.49 & \cellcolor{Gray}\textbf{52.50}\\
\hline
\end{tabular}}
\caption{1-shot and 5-shot class mIoU results on COCO-20$^i$ dataset for ResNet-50 as backbone, provided for 4 folds and the average. The best results are given in \textbf{boldface}. The \underline{underlined} results show the best performance excluding our method.}
\label{table:results_coco}
\end{table*}

\section{Experiments}

\subsection{Details}  

\noindent \textbf{Datasets.} The model is evaluated on two datasets which are commonly used in few-shot segmentation tasks. PASCAL-5$^{i}$ \cite{shaban} is the first dataset, containing 20 classes and it is a combination of PASCAL VOC 2012 \cite{voc} and the extended annotations obtained from \cite{sds}. The second dataset is COCO-20$^{i}$ \cite{coco}, which is generated from MSCOCO \cite{mscoco}. COCO-20$^{i}$ is more challenging when compared to PASCAL-5$^{i}$ as it consists of images belonging to 80 classes. The datasets are split into 4 folds containing equal number of classes in order to perform cross-validation while 1000 support and query pairs are randomly sampled for each fold. One of the folds is selected for evaluating the performance of the model on unseen classes while the rest of them are used as base classes for training the model. This procedure is repeated for all folds.

\noindent \textbf{Evaluation metrics.} In order to compare with previous studies on few-shot segmentation \cite{pgnet, pfenet, ntrenet, asnet, dpcn, bam}, class mean intersection-over-union is used as the evaluation metric, which is calculated as in \ref{eq:miou}, where $C$ is the number of classes in each fold.
\begin{equation}
\label{eq:miou}
    mIoU = \frac{1}{C} \sum_{i=1}^{C}IoU_i
\end{equation}

The foreground-background IoU (FB-IoU) is also calculated as an additional metric.

\noindent \textbf{Implementation details.} All experiments are conducted on PyTorch framework with NVIDIA RTX 2080Ti GPUs. As suggested in BAM \cite{bam}, there are two training stages, namely pre-training and meta-training. Pre-training stage is utilized for learning the base classes while ResNet-50 \cite{resnet} and VGG-16 \cite{vgg} are used as backbone for PASCAL-5$^i$ and only ResNet-50 \cite{resnet} is used as backbone for COCO-20$^i$. For PASCAL-5$^i$, PSPNet \cite{pspnet} is trained for 100 epochs as base learner with an initial learning rate of 2.5e-3. For the base learner on COCO-20$^i$, the model shared by the authors of \cite{bam} is used. In meta-training stage, PASCAL-5$^i$ and COCO-20$^i$ are trained for 200 and 50 epochs respectively while the learning rate is set to 5e-2. For both stages, SGD is utilized as optimizer. Random scaling, rotation, horizontal flip, cropping and Gaussian Blur is applied to images. The sizes of the enriched query features at the output of the improved decoder are set to 60, 30, 15, and 8, which makes N = 4 as suggested by \cite{pfenet}.

\begin{figure*}[t]
\centering
\resizebox{\linewidth}{!}{
\includegraphics[width=0.9\linewidth]{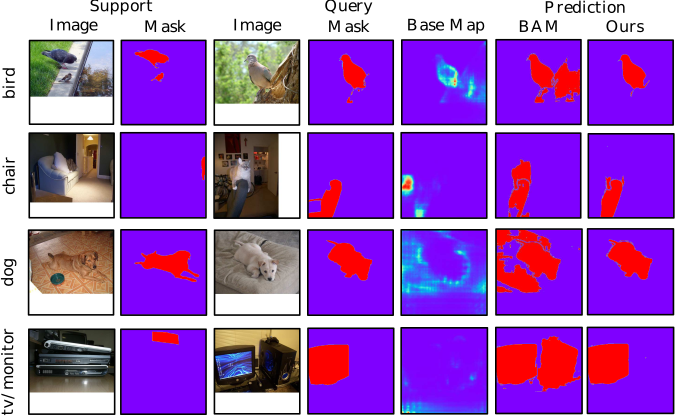}}
\caption{Qualitative 1-shot results on PASCAL-5$^i$ dataset for ResNet-50 backbone. Results for one novel class from each fold are provided in rows. First two columns contain image and mask for support while the following two columns contain image and ground truth for query. Fifth column shows the probability map for query obtained from base learner. Predictions are provided for BAM \cite{bam} and our method for comparison in the last two columns. (Best viewed in color)}
\label{fig:fqualitative}
\end{figure*}

\noindent \textbf{Generalized few-shot segmentation setting.} Our method is also evaluated in generalized few-shot segmentation setting, which is defined by \cite{bam}, where both pixels belonging to novel and base classes are detected. For this setting, novel pixels are predicted as \textit{novel} if their final foreground probabilities exceed a predefined threshold while the pixels predicted as \textit{base} should be assigned to one of the base classes. By this way, the pixels belonging to different base classes are distinguished while the rest of the pixels are classified as \textit{novel} or \textit{background}. This setting requires the calculation of mIoU on base and novel classes and also the combination of them, which are denoted by mIoU$_n$, mIoU$_b$ and mIoU$_a$ respectively.

\subsection{Results}

\subsubsection{Quantitative results} 

Table \ref{table:results_pascal} shows the performance comparison between BAM++ and other methods proposed for few-shot segmentation task using ResNet-50 and VGG-16. The mIoU results include 1-shot and 5-shot cases for PASCAL-5$^i$ dataset. BAM++ outperforms the existing methods for both settings. When VGG-16 is utilized as backbone, our method surpasses the state-of-the-art results by 1.61\% and 2.17\% for 1-shot and 5-shot settings, respectively. When it comes to the model with ResNet-50 as backbone, 0.78\% and 1.14\% performance gains are achieved for 1-shot and 5-shot settings. The results on COCO-5$^i$ dataset are provided in Table \ref{table:results_coco} for ResNet-50 as backbone only. BAM++ outperforms the best results by 0.93\% and 1.34\% under 1-shot and 5-shot settings, respectively. Comparison with state-of-the-art models regarding the FB-IoU scores is provided in Table \ref{table:fbiou} for both backbones on PASCAL-5$^i$ dataset. The results show that our method performs well in 1-shot setting while exceeds the best result by 0.66\% in 5-shot setting for ResNet-50. On the other hand, model with VGG-16 outperforms the previous state-of-the-art by 1.43\% and 1.42\% for 1-shot and 5-shot settings respectively.

\subsubsection{Qualitative results}

Qualitative results for PASCAL-5$^i$ dataset under 1-shot setting with ResNet-50 backbone are provided in Fig. \ref{fig:fqualitative}. The differences between our proposed architecture and BAM can be seen when the predicted masks are analyzed. The main advantage of our model is revealed in cases where there is another object adjacent to the novel target object. In such cases, models generally tend to entangle the objects. In Fig. \ref{fig:fqualitative}, it is seen that BAM predicts both the monitor and the computer as novel objects although there is only monitor in the support image. Since our model analyzes the features at different scales, it distinguishes the neighboring objects from each other well. Moreover, another faulty case is given in the third row, which is consistent with our hypothesis. Even though base learner discourages meta learner from non-novel regions, i.e. sofa, meta learner of BAM predicts these regions as novel. When ensembling the query predictions at different scales is introduced, such incorrect predictions are eliminated. As it can be seen in the predicted map of our method, only the regions belonging to the dog are considered as foreground. We deduce that ensembling at multi-scale ensures the model to focus on non-novel regions rather than the areas belonging to base classes.

\begin{table}[t]
    \centering
    \resizebox{\columnwidth}{!}{
    \begin{tabular}{|c|c|c|c|}
    \hline 
    Backbone & Method & 1-shot (\%) & 5-shot (\%)\\
    \hline
    \multirow{5}{*}{VGG-16} & PFENet (TPAMI'20) \cite{pfenet} & 72.00 & 72.30 \\
     & NTRENet (CVPR'22) \cite{ntrenet} & 73.10 & 74.20 \\
     & DPCN (CVPR'22) \cite{dpcn} & 73.70 & 77.20 \\
     & BAM (CVPR'22) \cite{bam} & \underline{77.26} & \underline{81.10} \\
     & \cellcolor{Gray}BAM++ (ours) & \cellcolor{Gray}\textbf{78.69} & \cellcolor{Gray}\textbf{82.52}\\
    \hline
    \multirow{6}{*}{ResNet-50} & PFENet (TPAMI'20) \cite{pfenet} & 73.30 & 73.90 \\
     & NTRENet (CVPR'22) \cite{ntrenet} & 77.00 & 78.40 \\
     & ASNet (CVPR'22) \cite{asnet} & 77.70 & 80.40 \\
     & DPCN (CVPR'22) \cite{dpcn} & 78.00 & 80.70 \\
     & BAM (CVPR'22) \cite{bam} & \underline{\textbf{79.71}} & \underline{82.18} \\
     & \cellcolor{Gray}BAM++ (ours) & \cellcolor{Gray}79.65 & \cellcolor{Gray}\textbf{82.84}\\
     \hline
    \end{tabular}
    }
    \caption{1-shot and 5-shot FB-IoU results on PASCAL-5$^i$ dataset for VGG-16 and ResNet-50 as backbone, provided as the average. The best results are given in \textbf{boldface}. The \underline{underlined} results show the best performance excluding our method.}
    \label{table:fbiou}
\end{table}

\subsubsection{Generalized few-shot segmentation results}

Our method surpasses BAM \cite{bam} in generalized few-shot segmentation setting for both backbones on PASCAL-5$^i$ dataset as shown in Table \ref{tab:generalizable_scenario}. The mIoU results validate the superiority of ensembling at multi-scale for both novel and base predictions.

\subsection{Ablation study}

Ablation study regarding the decision on how to include the inner losses for the multi-scale predictions are performed by considering the following cases: calculation of inner losses before and after the ensembling, without the ensembling, and after the ensembling only. The contributions of $\mathcal{L}_{meta}^{inner}$ in Eq. \ref{eq:loss_meta_inner} and $\mathcal{L}_{final}^{inner}$ in Eq. \ref{eq:loss_final_inner} on the final mIoU performance are investigated. Thus, we experimented with the cases where either $\mathcal{L}_{meta}^{inner}$ is inactive, $\mathcal{L}_{final}^{inner}$ is inactive, or both $\mathcal{L}_{meta}^{inner}$ and $\mathcal{L}_{final}^{inner}$ are active for the $\mathcal{L}_{final}^{total}$ calculation in Eq. \ref{eq:loss}. The results are obtained for PASCAL-5$^i$ dataset under 1-shot setting and provided in Table \ref{table:ablation_study}. Activating only $\mathcal{L}_{meta}^{inner}$ reaches an mIoU performance of 68.37\% while including $\mathcal{L}_{final}^{inner}$ alone obtains the performance of 68.45\%. The last row in Table \ref{table:ablation_study} indicates that when both $\mathcal{L}_{meta}^{inner}$ and $\mathcal{L}_{final}^{inner}$ are used, the highest performance is achieved, which is 68.59\%. As consequence, this ablation experiment validates our hypothesis, which emphasizes the weakness of the model implementing ensembling at single scale and the merits of the co-existence of $\mathcal{L}_{meta}^{inner}$ and $\mathcal{L}_{final}^{inner}$.

\begin{table}[t]
    \centering
    \resizebox{\linewidth}{!}{
    \begin{tabular}{|c|c|ccc|ccc|}
    \hline
    \multirow{2}{*}{Backbone} & \multirow{2}{*}{Method} & \multicolumn{3}{c|}{1-shot (\%)} & \multicolumn{3}{c|}{5-shot (\%)} \\
    \cline{3-8}
     &  & mIoU$_n$ & mIoU$_b$ & mIoU$_a$ & mIoU$_{n}$ & mIoU$_b$ & mIoU$_a$ \\
    \hline                          
    \multirow{2}{*}{VGG-16} & BAM \cite{bam} & 43.19 & 67.03 & 61.07 & 46.15 & 67.02 & 61.80 \\ 
                       & \cellcolor{Gray}BAM++ & \cellcolor{Gray}\textbf{43.94} & \cellcolor{Gray}\textbf{67.80} & \cellcolor{Gray}\textbf{61.83} & \cellcolor{Gray}\textbf{47.20} & \cellcolor{Gray}\textbf{67.80} & \cellcolor{Gray}\textbf{62.64} \\
    \hline
    \multirow{2}{*}{ResNet-50} & BAM \cite{bam} & 47.93 & 72.72 & 66.52 & 49.17 & 72.72 & 66.83 \\
                          & \cellcolor{Gray}BAM++ & \cellcolor{Gray}\textbf{49.98} & \cellcolor{Gray}\textbf{72.87} & \cellcolor{Gray}\textbf{67.15} & \cellcolor{Gray}\textbf{52.41} & \cellcolor{Gray}\textbf{72.87} & \cellcolor{Gray}\textbf{67.76} \\ 
    \hline
    \end{tabular}}
\caption{Generalized few-shot segmentation results on PASCAL-5$^i$ dataset for VGG-16 and ResNet-50 as backbone. The best results are given in \textbf{boldface}.}
\label{tab:generalizable_scenario}
\end{table}

\begin{table}[t]
    \centering
    \resizebox{0.6\linewidth}{!}{
    \begin{tabular}{|c|c|c|c|}
    \hline  
     Method & $\mathcal{L}_{inner}^{meta}$ & $\mathcal{L}_{inner}^{final}$ & mIoU (\%)\\
     \hline  
     BAM++ & \checkmark  & -           & 68.37 \\
     \hline  
     BAM++ & -            & \checkmark & 68.45 \\
     \hline  
     BAM++ & \checkmark  & \checkmark & 68.59 \\
     \hline  
     \end{tabular}}
     \caption{Ablation studies on inner losses for the multi-scale predictions regarding the ensembling with the base map under 1-shot setting for PASCAL-5$^i$. Results show the averaged mIoU over 4 folds.}
    \label{table:ablation_study}
\end{table}

\section{Conclusion}

We observed that although ensembling meta prediction with base prediction guides the model by making the meta learner cautious in the regions where objects from base classes exist, meta learner misclassifies non-novel regions by neglecting base learner. This situation arises as a consequence of ensembling the predictions at single-scale. Therefore, we proposed to perform ensembling for predictions at multi-scale as well as the final prediction. By this way, bias existing at non-novel regions is diminished. The experiments on PASCAL-5$^i$ and COCO-20$^i$ verifies our hypothesis and our model achieves new state-of-the-art on few-shot segmentation benchmark.


{\small
\bibliographystyle{ieee_fullname}
\bibliography{egbib}
}

\end{document}